%% file: main.tex
\documentclass[letterpaper, 10 pt, conference]{ieeeconf}  

\IEEEoverridecommandlockouts                              

\usepackage[pdftex]{graphicx}
\usepackage{epsfig} 
\usepackage{times} 
\usepackage{amsmath} 
\usepackage{amssymb}  
\usepackage{graphicx}

\usepackage{multicol}
\usepackage{hyperref}
\hypersetup{hidelinks}
\usepackage{xcolor}
\usepackage{makecell}
\usepackage{physics}
\usepackage{mathtools}
\usepackage{float}

\usepackage{booktabs}
\usepackage{wrapfig}
\usepackage{tabu}
\usepackage[ruled,vlined,algo2e]{algorithm2e}
\usepackage{algorithm}
\usepackage{algorithmic}
\usepackage{multirow}
\let\labelindent\relax
\usepackage{enumitem}

\usepackage[caption=false,font=footnotesize]{subfig}

\newcommand{\bfsection}[1]{\vspace*{0.1cm}\noindent\textbf{#1. }}

\makeatletter
\def\endthebibliography{%
  \def\@noitemerr{\@latex@warning{Empty `thebibliography' environment}}%
  \endlist
}
\makeatother

\usepackage{xspace}
\makeatletter
\DeclareRobustCommand\onedot{\futurelet\@let@token\@onedot}
\def\@onedot{\ifx\@let@token.\else.\null\fi\xspace}

\def\eg{\emph{e.g}\onedot} 
\def\ie{\emph{i.e}\onedot}

\makeatother

\title{\LARGE \bf
Loss Distillation via Gradient Matching for Point Cloud Completion with Weighted Chamfer Distance
}

\author{Fangzhou Lin$^{*1,3}$, Haotian Liu$^{*1}$, Haoying Zhou$^{*1}$, Songlin Hou$^{*2}$, \\Kazunori D Yamada$^{3}$, Gregory S. Fischer$^{1}$, Yanhua Li$^{4}$, Haichong K. Zhang$^{1}$, and Ziming Zhang$^{\dag,5}$  
\thanks{This work was supported by part of  NSF CCF-2006738, NIH DP5OD028162 and NSF AccelNet award OISE-1927275.}
\thanks{*These authors contributed equally to this work.}
\thanks{\dag Corresponding author.}
\thanks{$^{1}$Department of Robotics Engineering, Worcester Polytechnic Institute, Worcester, MA 01609, USA}%
\thanks{$^{2}$Dell Technologies, Hopkinton, MA 01748, USA}%
\thanks{$^{3}$Graduate School of Information Sciences, Tohoku University, Sendai, 980-8579, Japan}
\thanks{$^{4}$Department of Computer Science, Worcester Polytechnic Institute, Worcester, MA 01609, USA}
\thanks{$^{5}$Department of Electrical \& Computer Engineering, Worcester Polytechnic Institute, Worcester, MA 01609, USA 
({\tt\small zzhang15@wpi.edu})}
}

\begin{document}

\maketitle
\thispagestyle{empty}
\pagestyle{empty}

\input{sections/0_abstract} 

\input{sections/1_intro}
\input{sections/2_related_work}
\input{sections/3_approach}
\input{sections/4_experiments}
\input{sections/5_conclusion}

\clearpage

\bibliographystyle{IEEEtran}
\bibliography{references}

\end{document}

%% file: sections/0_abstract.tex
\begin{abstract}

3D point clouds enhanced the robot's ability to perceive the geometrical information of the environments, making it possible for many downstream tasks such as grasp pose detection and scene understanding. The performance of these tasks, though, heavily relies on the quality of data input, as incomplete can lead to poor results and failure cases.
Recent training loss functions designed for deep learning-based point cloud completion, such as Chamfer distance (CD) and its variants (\eg HyperCD \cite{lin2023hyperbolic}), imply a good gradient weighting scheme can significantly boost performance. However, these CD-based loss functions usually require data-related parameter tuning, which can be time-consuming for data-extensive tasks.
To address this issue, we aim to find a family of weighted training losses ({\em weighted CD}) that requires no parameter tuning. To this end, we propose a search scheme, {\em Loss Distillation via Gradient Matching}, to find good candidate loss functions by mimicking the learning behavior in backpropagation between HyperCD and weighted CD. Once this is done, we propose a novel bilevel optimization formula to train the backbone network based on the weighted CD loss. We observe that: (1) with proper weighted functions, the weighted CD can always achieve similar performance to HyperCD, and (2) the Landau weighted CD, namely {\em Landau CD}, can outperform HyperCD for point cloud completion and lead to new state-of-the-art results on several benchmark datasets. {\it Our demo code is available at \url{https://github.com/Zhang-VISLab/IROS2024-LossDistillationWeightedCD}.}



\end{abstract}


%% file: sections/1_intro.tex
\section{Introduction}
\label{sec:intro}



The applications of 3D point clouds widely expand to every corner of industrial and civilian areas like object recognition \cite{xie2018attentional,XU2022255}, mapping \cite{huitl2012tumindoor}, robotic grasping \cite{hu2024orbitgrasp, huang2023edge,huang2024imagination}, and pose estimation \cite{ten2017grasp}. However, because of occlusions, transparency, light reflections, or the limitation of the equipment's position and precision, the point clouds are usually sparse and incomplete \cite{leberl2010point}. 
To mitigate this issue, many learning-based point cloud completion methods \cite{guo2020deep} have been introduced, where supervised learning featured with a standard encoder-decoder architecture has emerged as the predominant choice for many researchers. These methods have achieved state-of-the-art performance on many benchmark datasets for point cloud completion \cite{pointtr, snowflakenet, zhou2022seedformer, wang2022pointattn}.

\begin{figure}[t]
    \centering
    \includegraphics[width=1\linewidth]{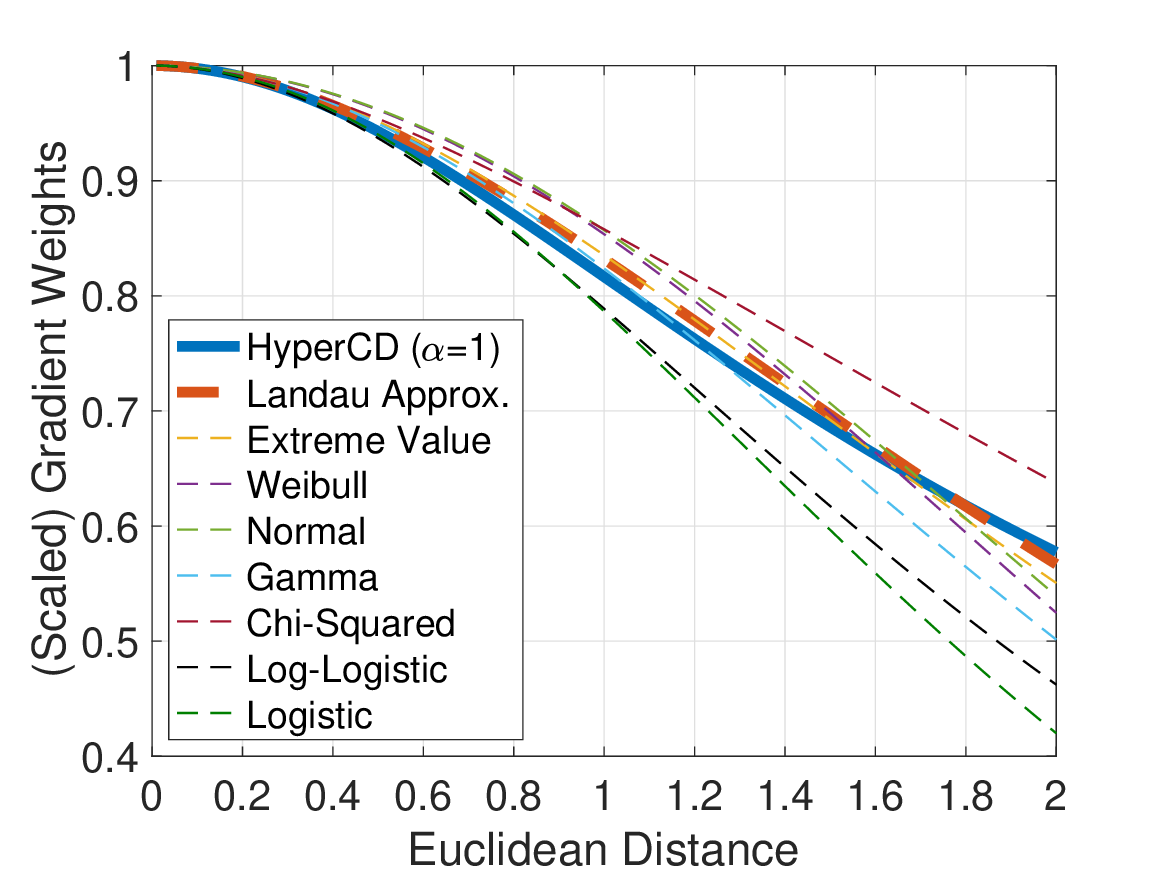}
	\caption{Illustration of distributions (with scaling and proper hyperparameters) that are similar to the gradient weighting distribution from HyperCD in backpropagation and can be taken as candidate weighting functions in weighted CD.}
    \label{fig:grad_weights}
\end{figure}

\subsection{Training Loss}
Chamfer distance (CD) serves as a popular training loss in point cloud completion for training neural networks such as SnowflakeNet \cite{snowflakenet} and PointAttN \cite{wang2022pointattn}. It evaluates the dissimilarity between any two point clouds by calculating the average distances of each point in one set to its nearest matching point in the other set. CD can faithfully reflect the global dissimilarity by treating the distances of all nearest-neighbor pairs between both sets with equal importance. However, it is not an ideal loss function solution for network training. The formation of CD works as the uniform distribution weight operation for paired distance, and thus, it is likely to be negatively affected by some outlier points. 
As the consequence, the sensitivity to outliers often results in a phenomenon where a considerable number of points from one set correspond to a single point in the other set, leading to the visual formation of small and dense clusters. This behavior can readily disrupt the commonly used assumption of uniform sampling from the underlying geometric surfaces, which is often used in the generation of point clouds. 
To mitigate these aforementioned problems in point cloud completion, several CD variant loss functions have been proposed: {\em Density-aware CD (DCD)} \cite{wu2021densityaware}, {\em InfoCD} \cite{lin2023infocd}, and {\em HyperCD} \cite{lin2023hyperbolic}.

\subsection{{Motivation}}
HyperCD utilizes the weighted gradients in backpropagation to update network weights. Though defined in a hyperbolic space, HyperCD equivalently assigns higher weights to the gradients from point pairs with smaller minimum Euclidean distances. The weighting scheme in HyperCD is only dependent on the Euclidean distance. This inspires us to address the following question: {\em Can we define a {\em weighted} CD to boost the performance of the vanilla CD?} To the best of our knowledge, we have not found any work on this topic in the literature on point cloud completion.

From a high-level understanding, we would like to propose a general method to efficiently learn loss functions for downstream tasks, as this is crucial for deep learning nowadays. Different from conventional hyperparameter tuning with scalars, we aim to explore functional spaces to search for good functions. Considering the specifications of point cloud completion, the sensitivity of CD to the outliers inspires us that the {\em distances in the metric as a training loss should be weighted in some form rather than uniform}. This provides us a good testbed to evaluate our high-level idea because currently, weighted CD is highly underexplored in this field. To address this issue, we borrow the idea from network distillation, where a simpler student network is trained to mimic the behavior of a more complicated teacher network. Rather than networks, we aim to learn weighted CD losses instead.


    

\subsection{Approach: Loss Distillation via Gradient Matching}
Loss functions guide the networks during training based on gradients through backpropagation, while gradients contain all the knowledge from the loss for training the networks. If we can reproduce the exact gradients in training, we can then reproduce the performance of a certain loss. Based on such considerations, we propose a family of training losses for point cloud completion using weighted CD to mimic the learning behavior of HyperCD by approximately matching the gradients. As illustrated in Fig. \ref{fig:grad_weights} (see more details in Sec. \ref{ssec:loss_distillation}), by taking the gradient weighting function in HyperCD as a reference, we can easily find some distributions as candidate weighting functions for weighted CD to approximate the reference curve, especially when the distance is small.

\subsection{Contributions}
We list our main contributions as follows:
\begin{itemize}
    
    \item We propose an efficient gradient-matching method for loss distillation to select candidate weighting functions for weighted CD from a pool of potential distributions.
    
    \item We demonstrate strong performance for point cloud completion based on weighted CD that can always be similar to our reference loss, even leading to state-of-the-art results on several benchmark datasets.
    
    
    \item {\em Our loss distillation method is so naive, yet effective and efficient, to determine good loss functions that all the calculations can be done using {\bf simple simulated data with mathematical derivations}.} It does provide us the solutions for our problems and potentially for other downstream tasks as well by matching reference gradients.
\end{itemize}

%% file: sections/2_related_work.tex
\section{Related Work}
\label{sec:rw}


\subsection{Distance Metrics for Point Cloud Completion} 
Distance in point clouds refers to a non-negative function that measures the dissimilarity among them \cite{dis_pc}. Considering the keen demand for high-density point cloud, the structures of point cloud completion networks have become increasingly complicated
\cite{ravi2020pytorch3d}. CD and its variants are extensively used in almost all recent learning-based methods for point cloud completion \cite{deng20193d, lyu2021conditional,zhang2022attention,tang2022lake}. 
\subsection{Knowledge Distillation (KD)}
Generally, KD \cite{knowledge_distilling} refers to a model compression method in machine learning, where a smaller, more compact neural network (\ie student model) is trained to replicate the behavior of a larger, more complex network (\ie teacher model) \cite{kd_teacher_student}. The teacher model is used to produce the outputs of knowledge, while the student model tries to learn such knowledge by mimicking the outputs. Some nice survey papers on this topic can be found in \cite{gou2021knowledge, wang2021knowledge, alkhulaifi2021knowledge}. 
\subsection{Weighted Chamfer Distance}
In 2D image processing, weighted distances have become notable in generating distance maps from point lattice \cite{WD_PointLattice} and image segmentation \cite{ImageSeg_via_WD}. In the distance map generation task, this methodology facilitates the computation of rotation-invariant distances through optimal weighting, particularly in face-centered cubic \cite{fcc} and body-centered cubic lattice structures \cite{WD_PointLattice, bcc}. In 3D point cloud applications, the weighted CD emerges as a pivotal loss function and metric \cite{loss_object_func, neighbor_loss, deep_p_dis}. However, to the best of our knowledge, in point cloud completion we do not find any reference based on weighted CD. 


%% file: sections/3_approach.tex
\section{Our Approach}
\label{sec:approach}




\subsection{Chamfer Distance (CD)} 
We denote $(x_i,y_i)$ as the $i$-th point cloud pair, $x_i=\{x_{ij}\}$ and $y_i=\{y_{ik}\}$ as two sets of 3D points, and $d(\cdot,\cdot)$ as a certain distance metric. Then the CD loss for point clouds can be defined as follows:
\begin{align}\label{eqn:CD}
    & \hspace{-2mm} D_{CD}(x_i, y_i) \nonumber \\ & \hspace{-2mm} = \frac{1}{|x_i|}\sum_{j=1}^{|x_i|}\min_k d(x_{ij}, y_{ik}) + \frac{1}{|y_i|}\sum_{k=1}^{|y_i|}\min_j d(x_{ij}, y_{ik}),
\end{align}
where $|\cdot|$ denotes the cardinality of a set. Note that for point cloud completion, function $d$ is usually defined in Euclidean space, referring to 
\begin{align}
    d(x_{ij}, y_{ik})=\left\{
    \begin{array}{ll}
        \|x_{ij} - y_{ik}\| & \mbox{as {\em L1-distance}} \\
        \|x_{ij} - y_{ik}\|^2 & \mbox{as {\em L2-distance}}
    \end{array}
    \right.
\end{align}
where $\|\cdot\|$ denotes the $\ell_2$ norm of a vector.

\subsection{Hyperbolic Chamfer Distance (HyperCD)}
Based on Eq. \ref{eqn:CD}, HyperCD defines the function $d$ in a hyperbolic space as follows:

\begin{align}
    d(x_{ij},y_{ik}) = arccosh\left(1+\alpha\|x_{ij}-y_{ik}\|^2\right), \alpha>0.
    \label{eqn:hypercd}
\end{align}

\begin{figure*}[ht]
    \hfill
	\begin{minipage}[b]{0.325\linewidth}
		\begin{center}
			\centerline{\includegraphics[width=\linewidth, keepaspectratio,]{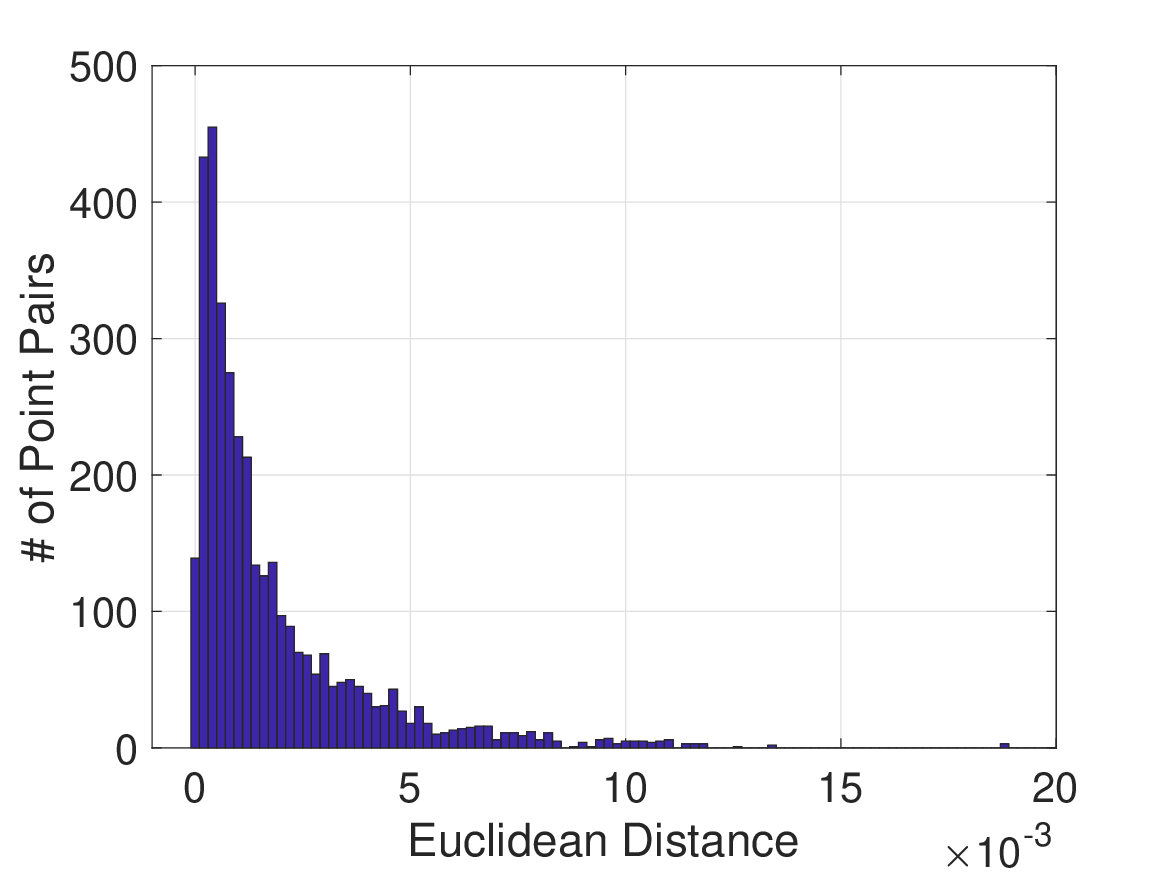}}
			\centerline{(a)}
		\end{center}
	\end{minipage}
	\hfill
	\begin{minipage}[b]{0.325\linewidth}
		\begin{center}
			\centerline{\includegraphics[width=\linewidth,keepaspectratio]{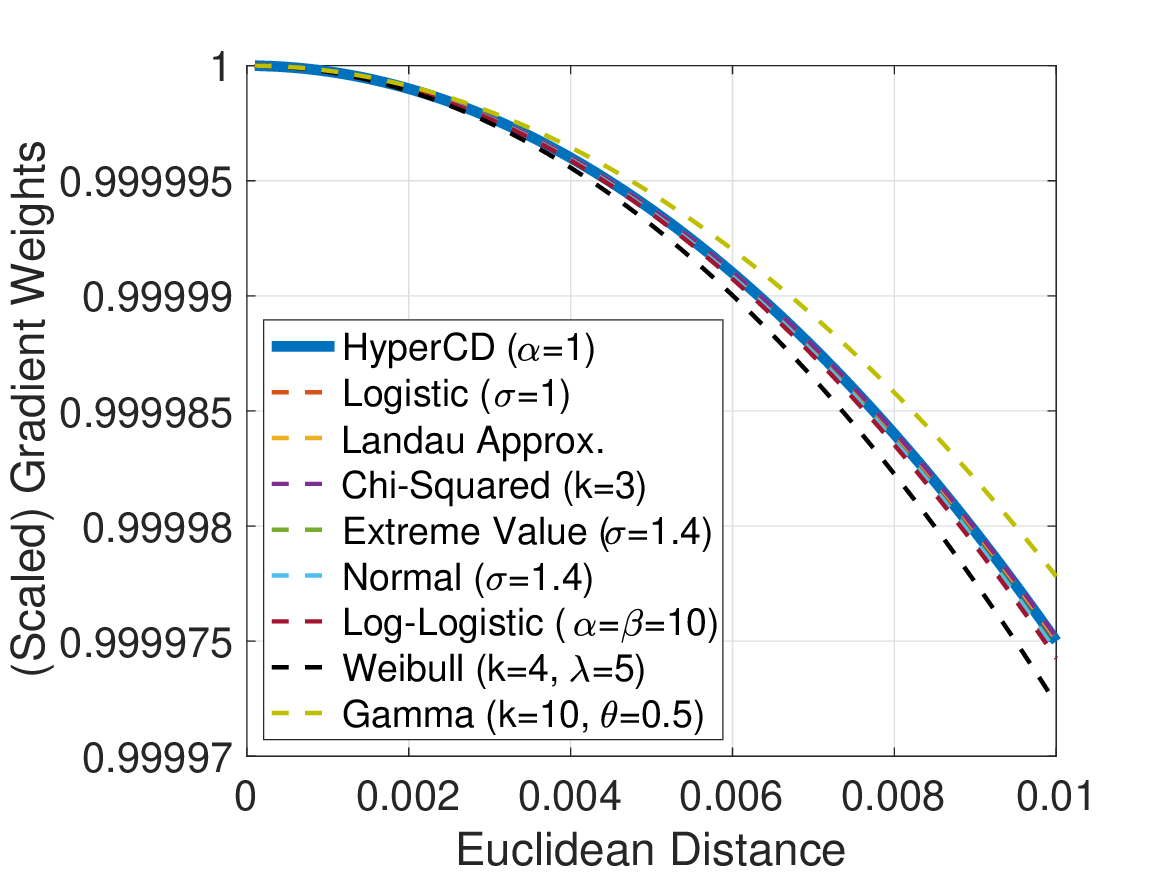}}
			\centerline{(b)}
		\end{center}
	\end{minipage}
 \hfill
	\begin{minipage}[b]{0.325\linewidth}
		\begin{center}
			\centerline{\includegraphics[width=\linewidth,keepaspectratio]{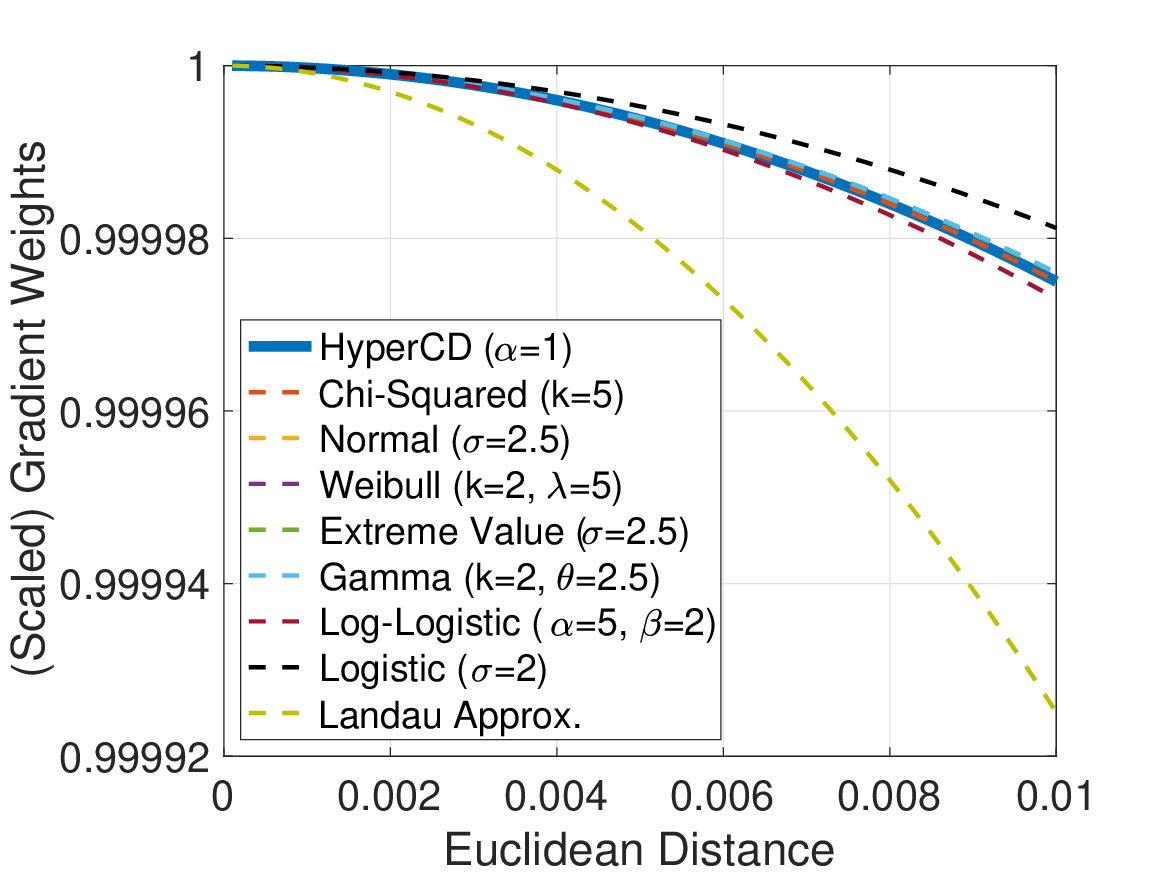}}
			\centerline{(c)}
		\end{center}
	\end{minipage}
	\vspace{-5mm}
    \caption{Illustration of {\bf (a)} reference distance distribution from HyperCD, and {\bf (b-c)} curve fitting using different approximations of $z^{(W)}$.}
\label{fig:opt_grad_matching}
\end{figure*}

\subsection{Loss Distillation with Weighted CD}\label{ssec:loss_distillation}

\bfsection{Weighted Chamfer Distance}
In this paper, we propose the following formula for our weighted CD:
\begin{align}\label{eqn:weightedCD}
    & D_{W}(x_i,y_i) = \frac{1}{|x_i|}\sum_{j=1}^{|x_i|} f(\Tilde{d}_{ijk}) \Tilde{d}_{ijk} + \frac{1}{|y_i|}\sum_{k=1}^{|y_i|} f(\Tilde{d}_{ikj}) \Tilde{d}_{ikj} \nonumber \\
    & \mbox{s.t.} \, \Tilde{d}_{ijk} = \min_k d(x_{ij}, y_{ik}), \Tilde{d}_{ikj} = \min_j d(x_{ij}, y_{ik}).
\end{align}
Clearly, the vanilla CD is a special case of our new formula with $f(\Tilde{d}_{ijk}) = f(\Tilde{d}_{ikj}) = 1, \forall \Tilde{d}_{ijk}, \Tilde{d}_{ikj} \geq 0$.

\bfsection{Loss Distillation via Gradient Matching}
For point cloud completion, let us denote $(x_i,y_i)$ as the output from the network and the ground-truth point cloud, respectively. Precisely, we denote $x_i = h(\Tilde{x}_i; \omega)$ where function $h$ is presented by the network with parameters $\omega$ and $\Tilde{x}_i$ is an incomplete point cloud as the input. Therefore, each point $x_{ij}$ is also a function of $\omega$, and so are $\Tilde{d}_{ijk}$ and $\Tilde{d}_{ikj}$.

Recall that the goal of gradient matching is to develop effective losses based on weighted CD by mimicking the learning behavior of HyperCD. To simplify our explanation of gradient matching for loss distillation, we denote $g_{ijk}^{(H)} = arccosh\left(1+\alpha\Tilde{d}_{ijk}^2\right), g_{ijk}^{(W)} = f(\Tilde{d}_{ijk}) \Tilde{d}_{ijk}$. 
To match a pair of gradients from both losses, we propose minimizing their difference as follows:

\begin{equation}
    \scriptsize
    \begin{split}
        & \left\|\frac{\partial D_H}{\partial \omega} -\frac{\partial D_W}{\partial \omega} \right\| \nonumber \\
    & \leq \frac{1}{|x_i|}\left\|\sum_j\left(\frac{\partial g_{ijk}^{(H)}}{\partial\omega} - \frac{\partial g_{ijk}^{(W)}}{\partial\omega}\right)\right\| + \frac{1}{|y_i|}\left\|\sum_k\left(\frac{\partial g_{ikj}^{(H)}}{\partial\omega} - \frac{\partial g_{ikj}^{(W)}}{\partial\omega}\right)\right\| \nonumber \\
    &
    \leq \frac{1}{|x_i|}\sum_j \left\|z_{ijk}^{(H)} - z_{ijk}^{(W)}\right\|\left\|\frac{\partial \Tilde{d}_{ijk}}{\partial\omega}\right\| + \frac{1}{|y_i|}\sum_k \left\|z_{ikj}^{(H)} - z_{ikj}^{(W)}\right\|\left\|\frac{\partial \Tilde{d}_{ikj}}{\partial\omega}\right\| 
    \end{split}
    \label{eqn:grad_match}
\end{equation}

where $z_{ijk}^{(H)} = \frac{2\alpha \Tilde{d}_{ijk}}{\sqrt{\left(1+\alpha \Tilde{d}_{ijk}^2\right)^2-1}}, z_{ijk}^{(W)} = f'(\Tilde{d}_{ijk}) \Tilde{d}_{ijk} + f(\Tilde{d}_{ijk})$ are {\em gradient weights} for the HyperCD loss, $D_H$, and the weighted CD loss, respectively ({\it resp.} $z_{ikj}^{(H)}, z_{ikj}^{(W)}$), and $f'$ denotes the derivative of function $f$. 

Minimizing the LHS of Eq. \ref{eqn:grad_match} is very challenging, because we do not have prior knowledge about the network and data. To get rid of the effects of such unknown information in learning, we instead try to minimize the RHS of Eq. \ref{eqn:grad_match} with the following assumptions on
\begin{itemize}[nosep, leftmargin=*]
    \item {\em Network:} All the gradients can be upper-bounded.
    \item {\em Data:} $|x_i|, |y_i|$ are sufficiently large so that the distributions of $\Tilde{d}_{ijk}, \Tilde{d}_{ikj}$ follow the reference distance distribution from HyperCD.
\end{itemize}
In point cloud completion, both assumptions can hold easily. Finally, due to the symmetry of Eq. \ref{eqn:grad_match}, we propose the following minimization problem for loss distillation:
\begin{align}\label{eqn:loss_distillation}
    & \min_{f\in\mathcal{F}}\mathbb{E}_{\Tilde{d}\sim\Tilde{\mathcal{D}}}\left\|z^{(H)}(\Tilde{d}) - z^{(W)}(\Tilde{d})\right\| \nonumber \\
    \approx & \min_{f\in\mathcal{F}}\sum_{\Tilde{d}} p(\Tilde{d})\left\|z^{(H)}(\Tilde{d}) - z^{(W)}(\Tilde{d})\right\|,
\end{align}
where $z^{(H)}(\Tilde{d}) = \frac{2\alpha \Tilde{d}}{\sqrt{\left(1+\alpha \Tilde{d}^2\right)^2-1}}, z^{(W)}(\Tilde{d}) = f'(\Tilde{d}) \Tilde{d} + f(\Tilde{d})$,  $\mathcal{F}$ denotes the feasible space for $f$, $\Tilde{D}$ denotes the reference distance distribution, and $\mathbb{E}$ denotes the expectation operator. Note that when Eq. \ref{eqn:loss_distillation} reaches 0 , it will guarantee to recover the performance of HyperCD using weighted CD.

\begin{table}[htbp]
\tabulinesep=1.2mm
\caption{Distributions as weighting functions in weighted CD.}
\resizebox{\columnwidth}{!}{
\begin{tabu}{l|l|l|l}
\hline
\textbf{Distribution} & \textbf{Params} & \textbf{Mode} $m$ & \textbf{PDF} \\ \hline
Chi-Squared            & $k$     &    $\max(k-2, 0)$      &  $\displaystyle \frac{1}{2^{k/2}\Gamma(k/2)} x^{k/2 - 1} e^{-x/2}$           \\ \hline
Extreme Value         & $\beta$         &    0   & $\displaystyle \frac{1}{\beta} e^{-(z + e^{-z})}, z = \frac{x}{\beta}$             \\ \hline
Weibull               & $k, \lambda$     &   $\begin{cases} 
\lambda \left(\frac{k-1}{k}\right)^{1/k}, & k > 1, \\
0, & k \leq 1.
\end{cases}$        & $\displaystyle \frac{k}{\lambda} \left(\frac{x}{\lambda}\right)^{k-1} e^{-(x/\lambda)^k}$             \\ \hline
Log-Logistic          & $\alpha, \beta$     &      $\begin{cases} 
\alpha \left(\frac{\beta - 1}{\beta + 1}\right)^{1/\beta}, & \text{if } \beta > 1, \\
0, & \text{otherwise}
\end{cases}$     & $\displaystyle \frac{\beta}{x} \left(1 + \left(\frac{x}{\alpha}\right)^{-\beta}\right)^{-1-\beta}$             \\ \hline
Gamma                 & $\alpha, \beta$        &   $\begin{cases} 
\frac{\alpha - 1}{\beta}, & \text{for } \alpha \geq 1, \\
0, & \text{for } \alpha < 1
\end{cases}$     & $\displaystyle \frac{\beta^\alpha x^{\alpha-1} e^{-\beta x}}{\Gamma(\alpha)}$             \\ \hline
Logistic              & $\sigma$        &    0    & $\displaystyle \frac{e^{-x/\sigma}}{\sigma(1 + e^{-x/\sigma})^2}$             \\ \hline
Normal                & $\sigma$        &    0    & $\displaystyle \frac{1}{\sigma \sqrt{2\pi}} \exp\left(-\frac{x^2}{2\sigma^2}\right)$             \\ \hline
Landau Approx.                &   -   &    0    & $\displaystyle \frac{1}{\sqrt{2\pi}}\exp{\left(- \frac{x + e^{-x}}{2}\right)}$             \\ \hline
\end{tabu}}
\label{tab:dists}
\end{table}

\subsection{Optimization}
To solve Eq. \ref{eqn:loss_distillation}, we first specify the notations in our implementation as follows:
\begin{itemize}[nosep, leftmargin=*]
    \item {\em Feasible space $\mathcal{F}$:} Table \ref{tab:dists} lists some well-known distributions that we tested as the weighting functions for weighted CD. Each distribution defines an $\mathcal{F}$, and we try to learn its parameters, if exist, to determine $f$. 

    \item {\em Distance $\Tilde{d}$:} Recall that $z^{(H)}$ in HyperCD is monotonically decreasing and $\Tilde{d}=0$ reaches the maximum. To mimic this behavior, we only consider the partial distributions beyond their modes. Correspondingly, the input data for each distribution is its mode, $m$, plus distance $\Tilde{d}$.

    \item {\em Reference distance distribution $\Tilde{\mathcal{D}}$ and samples $\Tilde{d}$:} Fig.~\ref{fig:opt_grad_matching} (a) illustrates the distance distribution from HyperCD, which is used as $\Tilde{\mathcal{D}}$ in our implementation. As we see, about 99\% of point pairs, \ie $\Tilde{d}$, fall into $[0,0.01]$. Therefore, to optimize Eq. \ref{eqn:loss_distillation} efficiently, we uniformly sample $\Tilde{d}$ from $[0,0.01]$ with step 2e-4, and the corresponding probabilities are sampled from Fig. \ref{fig:opt_grad_matching} (a). 

    \item {\em Approximation of $z^{(W)}$:} The exact computation of $f'$ in $z^{(W)}$ causes trouble, even if we may know its function (\eg for some functions we may not have their analytical solutions of their gradients). To address this issue, we propose the following two ways: (1) $z^{(W)}(\Tilde{d}) \approx f(m+\Tilde{d})$, because $\Tilde{d}\in[0,0.01]$ is very small and thus the calculation of $z^{(W)}$ may be dominated by the second term; or (2) substituting $f'(\Tilde{d})\approx\frac{1}{\Delta\Tilde{d}}(f(m+\Tilde{d}+\Delta\Tilde{d}) - f(m+\Tilde{d}))$ into $z^{(W)}$ with small value $\Delta\Tilde{d}\geq0$. Based on these two strategies, we plot the curves in Fig. \ref{fig:opt_grad_matching} (b-c), respectively, where all the curves are rescaled by the maximum values and ordered by the minimum of Eq. \ref{eqn:loss_distillation}. The optimal parameters are determined using a grid search for simplicity and efficiency. As we see, all eight distributions can well fit the reference curve from HyperCD, and the parameters listed in the figure are used to report the performance of weighted CD in our experiments. Finally, we list our gradient matching algorithm in Alg.~\ref{alg:grad_matching}.
\end{itemize}

\begin{algorithm}[htbp]
    \small
    \SetAlgoLined
    \SetKwInOut{Input}{Input}\SetKwInOut{Output}{Output}
    \Input{a PDF $f$ with parameters $\mathcal{A}$, $z^{(H)}$, $\{(\Tilde{d}, p(\Tilde{d}))\}$}
    \Output{$\mathcal{A}$}
    \BlankLine
    Discretize the parameter space into $\{\mathcal{A}_i\}$ for grid search;

    Compute the mode $m_i$ for each $\mathcal{A}_i$ used in $z^{(W)}$;

    $\mathcal{A}^* = argmin_{\{\mathcal{A}_i\}} \sum_{\Tilde{d}} p(\Tilde{d})\left\|z^{(H)}(\Tilde{d}) - z^{(W)}(\Tilde{d})\right\|$;
    
    \Return $\mathcal{A}\leftarrow\mathcal{A}^*$
    \caption{Loss Distillation via Gradient Matching}\label{alg:grad_matching}
\end{algorithm}

\begin{algorithm}[htbp]
    \small
    \SetAlgoLined
    \SetKwInOut{Input}{Input}\SetKwInOut{Output}{Output}
    \Input{a weighting function $f$, a network $h$ with learnable parameters $\omega$, training data $\{(\Tilde{x}_i, y_i)\}$}
    \Output{$\omega$}
    \BlankLine
    \Repeat{converges}
    {
    Pick a sample $(\Tilde{x}_i, y_i)$ uniformly at random;

    Compute $\Tilde{d}_{ijk}, \forall j$ in $\Tilde{x}_i$ and $\Tilde{d}_{ikj}, \forall k$ in $y_i$;

    Compute the weighted CD loss based on Eq. \ref{eqn:weightedCD};

    Update the parameters $\omega$ using the gradient of the loss;
    
    }
    \Return $\omega$
    \caption{Point Cloud Completion with Weighted CD}\label{alg:bilevel_opt}
\end{algorithm}




\begin{figure*}[htbp]
    \centering
    \includegraphics[width=0.85\linewidth]{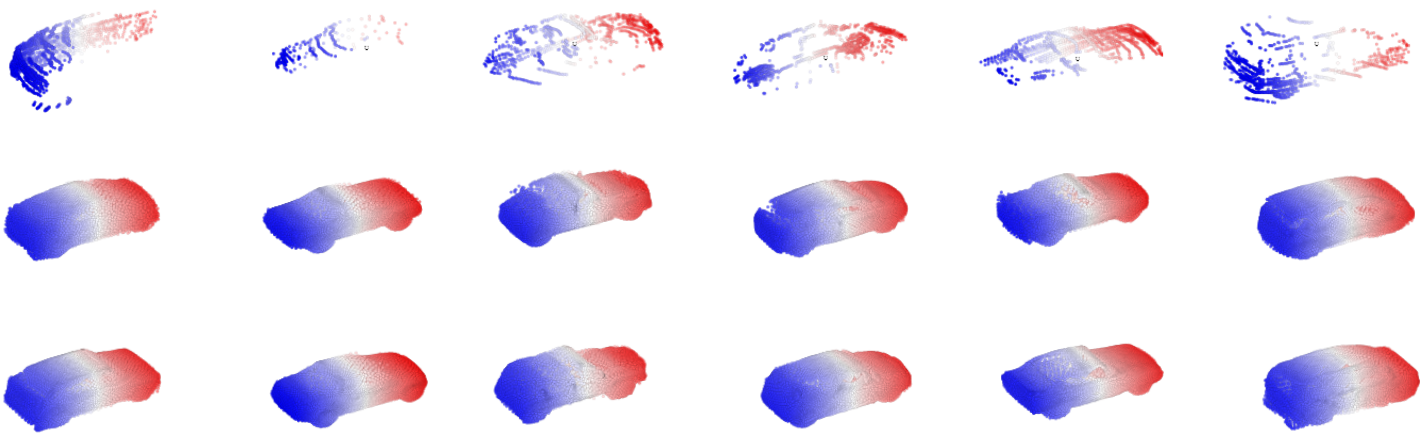}
    \caption{Visualization of the real-world(KITTI) benchmark (\textbf{Row 1:} sparse input, \textbf{Row 2:} HyperCD, \textbf{Row 3:} LandauCD).}
    \label{fig:kitti}
\end{figure*}


\subsection{Bilevel Optimization with Weighted CD}

Once we choose the weighting function in weighted CD, we propose optimizing the following optimization problem for point cloud completion, given training samples $\{(\Tilde{x}_i, y_i)\}$:
\begin{align}
    & \min_{\omega}\sum_i\left[\frac{1}{|x_i|}\sum_{j=1}^{|x_i|} f(\Tilde{d}_{ijk}) \Tilde{d}_{ijk} + \frac{1}{|y_i|}\sum_{k=1}^{|y_i|} f(\Tilde{d}_{ikj}) \Tilde{d}_{ikj}\right] \nonumber \\
    & \mbox{s.t.} \, x_i = h(\Tilde{x}_i; \omega) = \{x_{ij}\}, \forall i, \nonumber \\
    & \hspace{5mm} \Tilde{d}_{ijk} = \min_k d(x_{ij}, y_{ik}), \Tilde{d}_{ikj} = \min_j d(x_{ij}, y_{ik}).
\end{align}

Essentially, this defines a bilevel optimization problem that can be solvable using the iterative differentiation algorithm \cite{ji2021bilevel}, leading to our algorithm in Alg. \ref{alg:bilevel_opt}. In fact, the learning algorithms for both HyperCD follow the same bilevel optimization strategy as ours. 

%% file: sections/4_experiments.tex
\section{Experiments}
\label{sec:exp}

\subsection{Datasets and Network Backbones Description} 
\bfsection{Datasets}In our experiments we utilize PCN \cite{pcn}, ShapeNet-55/34 \cite{pointtr}, ShapeNet-Part \cite{yi2016scalable}, KITTI \cite{geiger2013vision}, and SVR ShapeNet \cite{xiang2022snowflake}. Please refer to HyperCD for the dataset details of PCN, ShapeNet-55/34, and ShapeNet-Part. KITTI is composed of a sequence of real-world Velodyne LiDAR scans, also derived from the PCN dataset \cite{pcn}. For each frame, the car objects are extracted according to the 3D bounding boxes, which results in 2,401 partial point clouds. The partial point clouds in KITTI are highly sparse and do not have complete point clouds as ground truth. SVR ShapeNet is also generated from the synthetic ShapeNet \cite{chang2015shapenet} which contains 43,783 shapes from 13 categories. The dataset is split into training, testing, and validation sets by the ratios of 70\%, 20\%, and 10\%, respectively.

\bfsection{Network Backbonses} We compare our method using 7 different backbone networks, \ie FoldingNet \cite{yang2018foldingnet}, PMP-Net \cite{wen2021pmp}, PoinTr \cite{pointtr}, SnowflakeNet \cite{snowflakenet}, CP-Net \cite{lin2022cosmos}, PointAttN \cite{wang2022pointattn} and SeedFormer \cite{zhou2022seedformer}, by replacing the CD loss with our weighted CD losses wherever it occurs.
\subsection{Hyperparameters}
The hyperparameters in the weighting functions are selected from the candidate functions shown in Fig. \ref{fig:opt_grad_matching} (b-c) that achieve better performance. Except that the learning rates are tuned slightly, the training hyperparameters such as batch sizes and balance factors in the original losses are kept the same as HyperCD. 

\subsection{Evaluation}
Following the literature, we evaluate the best performance of all the methods using vanilla CD (lower is better). We also use F1-Score@1\% \cite{tatarchenko2019single} (higher is better) to evaluate the performance on ShapeNet-55/43. For KITTI, we use Fidelity and MMD metrics \cite{pointtr}. For better comparison, we cite the original results of some other methods on PCN, ShapeNet-55, and KITTI. In each table of the results, the top-performing results are highlighted in red, while the second-highest ones are marked in blue.

\begin{figure}[htbp]
    \centering
    \includegraphics[width=0.985\linewidth]{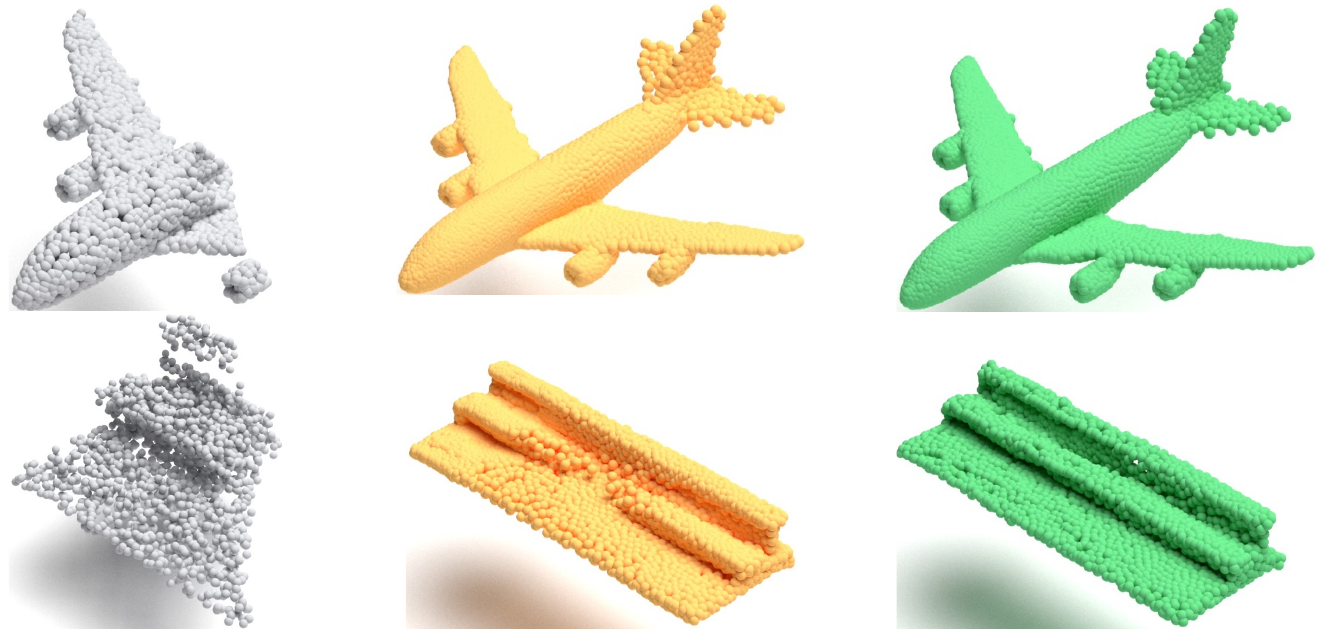}
    \caption{Visualization of ShapeNet-55 benchmark. Gray represents the partial input. Yellow represents HyperCD. Green represents Landau CD.}
    \label{fig:shapnet55}
\end{figure}

\begin{table}[htb]
\centering
    \caption{CP-Net comparison results on ShapeNet-Part with different losses. In the sequel, we color the best performance with {\em red}, and second best with {\em blue}.}
    \label{table:Shapenet-Part_analysis_2}
    \footnotesize
        \begin{tabular}{c|c|ccc}
    		\toprule
    		Loss Function & Loss Params.& L2-CD$\times10^3$ \\
    		\midrule
    		CD & $\backslash$ &4.16 \\
    		EMD & $\backslash$ &15.38 \\
            Truncated CD & thd=0.2 &4.72 \\
    		DCD \cite{wu2021densityaware} & $\alpha$ = 40, $\gamma$=0.5& 5.74 \\     
            HyperCD \cite{lin2023hyperbolic} &  $\alpha$=1&\textcolor{blue}{4.03} \\
            \midrule

            {\bf Weibull CD} & $k$=2, $\lambda$=5 & 4.19\\
            {\bf Normal CD} & $\sigma$=1.4 &4.17\\
            {\bf Logistic CD} & $\sigma$=1 & 4.14\\  
            {\bf Log-Logistic CD} & $\alpha$=5, $\beta$=2 &  4.12\\ 


            {\bf Extreme-Value CD} & $\sigma$=1.4 & 4.08\\
            {\bf Chi-Squared CD} & $k$=3 & 4.07 \\ 
            {\bf Gamma CD} & $k$=2, $\theta$=2.5 & \textcolor{blue}{4.03}\\
            {\bf Landau CD} &$\backslash$ & 
            \textcolor{red}{ 4.00$\pm$0.005} \\
    		\bottomrule
	    \end{tabular}
     
\end{table}

\subsection{Ablation Study}
For this purpose, we use CP-Net as the backbone network and train it on ShapeNet-Part with different losses. We refer to our different weighted CDs based on the names of the distributions as weighting functions. For instance, we call a weighted CD {\bf Landau CD} if the weighting function follows the Landau approximation distribution. 

\subsection{Performance}
\begin{figure*}[ht]
	\begin{center}
		\includegraphics[width=0.9\linewidth]{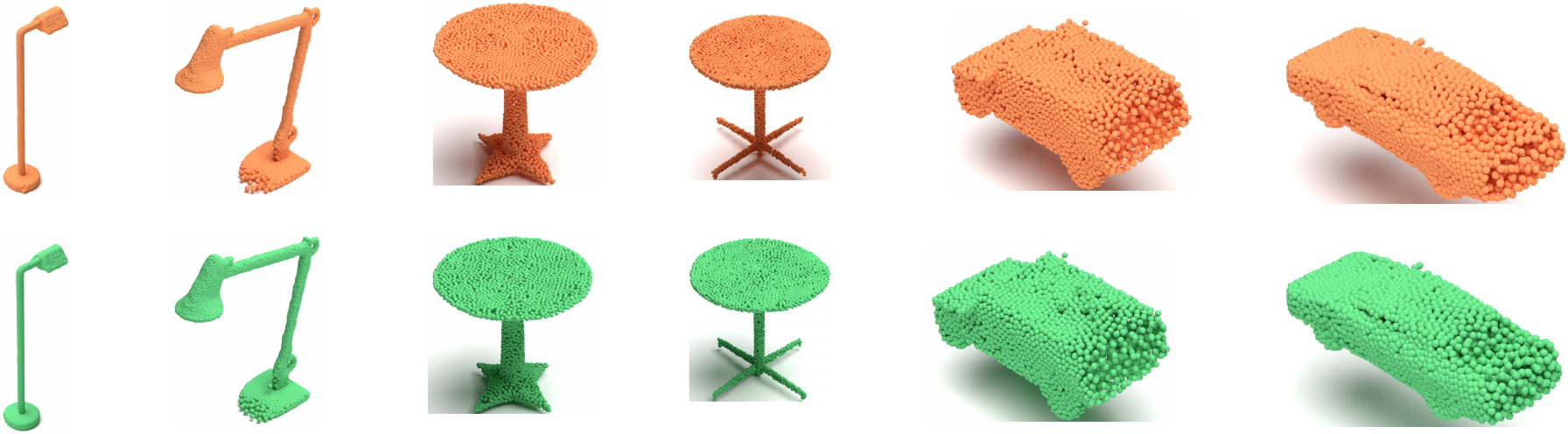}
	\end{center}
    \vspace{-3mm}
	\caption{{Visual comparison on Shapenet-55. {\bf Row-1:} Seedformer with HyperCD.  {\bf Row-2:} Seedformer with Landau CD.}}
	\label{fig:shapnet55_large}
\end{figure*}

Table~\ref{table:Shapenet-Part_analysis_2} summarizes our comparison results where we report the best performance for all the methods (our loss parameters are chosen from Fig. \ref{fig:opt_grad_matching} (b-c) through gradient matching). As we see here, 6 out of 8 weighted CD losses perform better than CD, and 4 of them perform similarly to HyperCD (within the difference of $\pm0.05$). Surprisingly our Landau CD even beats the state-of-the-art. Notice that the performance ranking of weighted CD losses is not consistent with the function matching ranking in Fig. \ref{fig:opt_grad_matching} (b-c), indicating that the selected weighting functions have to be tested with the weighted CD losses. Overall, such results demonstrate that our weighted CD can mimic the learning behavior of HyperCD, with proper weighting functions and parameters, and have great potential for boosting CD performance significantly. Fig. \ref{fig:shapnet55_large} visualizes some results between HyperCD and our Landau CD, where the differences are marginal. This, again, demonstrates our success in mimicking HyperCD learning behavior.

\subsection{Convergence}
\begin{figure}[htb]
	\begin{center}
		\includegraphics[width=1\linewidth]{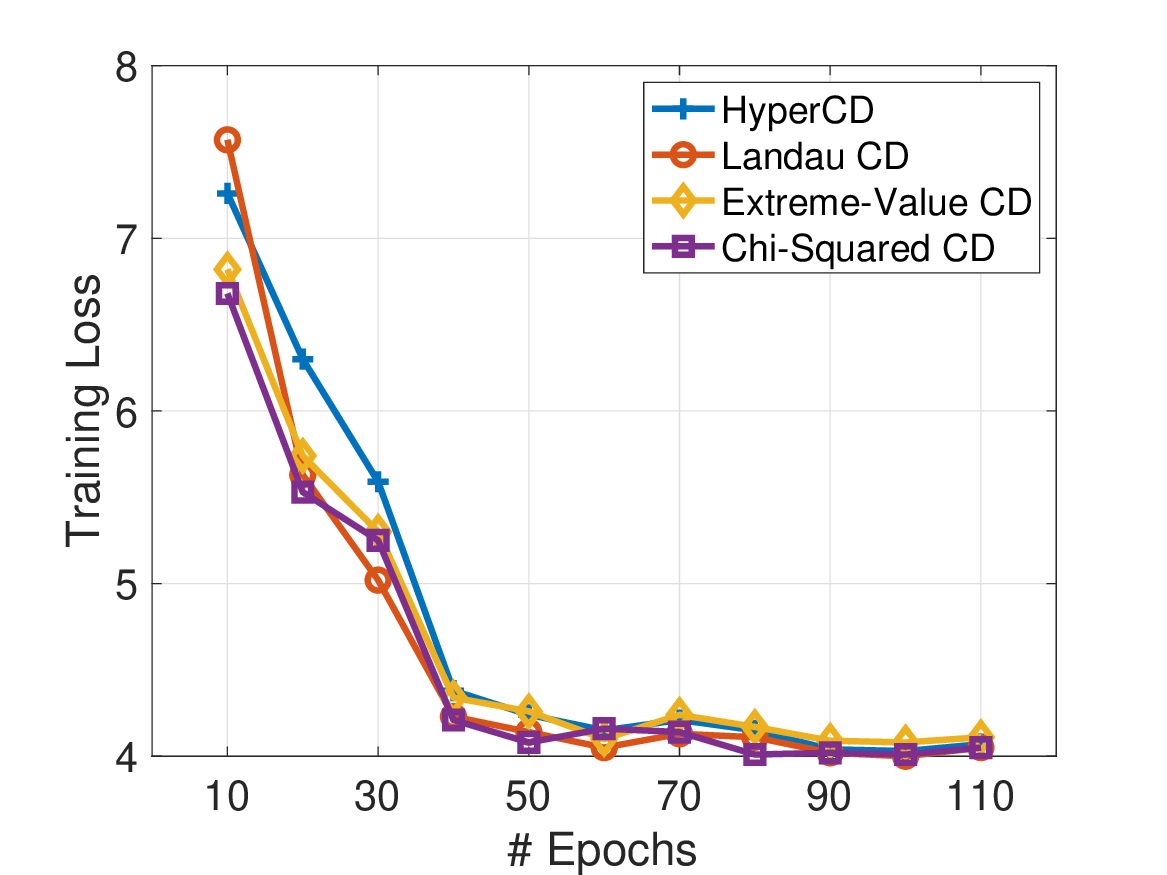}
	\end{center}
    \vspace{-3mm}
	\caption{Loss convergence of CP-Net on ShapeNet-Part.}
	\label{fig:convergence}
\end{figure}

Fig.~\ref{fig:convergence} provides a direct visual juxtaposition of the convergence trends, revealing that our weighted CD losses exhibit a more rapid convergence compared to HyperCD. Notably, during the initial 50 epochs, the curves of weighted CD loss functions consistently remain lower than the ones of HyperCD, and eventually, all the curves converge to a similar loss, leading to similar performance as well. Such convergence behavior of our weighted CD also demonstrates the success of our loss distillation method. 

\subsection{Computational Time}
The computation of our weighted CD is dominated by the weighting functions. However, in practice, we do not observe a significant difference between ours and HyperCD or InfoCD. For instance, using CP-Net on ShapeNet-Part, during training the computational time of CD, HyperCD, InfoCD and Landau CD per iteration is 0.4239$\pm$0.0019, 0.4298$\pm$0.0014, 0.4498$\pm$0.0030, and 0.4341$\pm$0.0026, respectively.

\subsection{More analysis with our insight}
\begin{figure}[t]
	\begin{center}
		\includegraphics[width=1\linewidth]{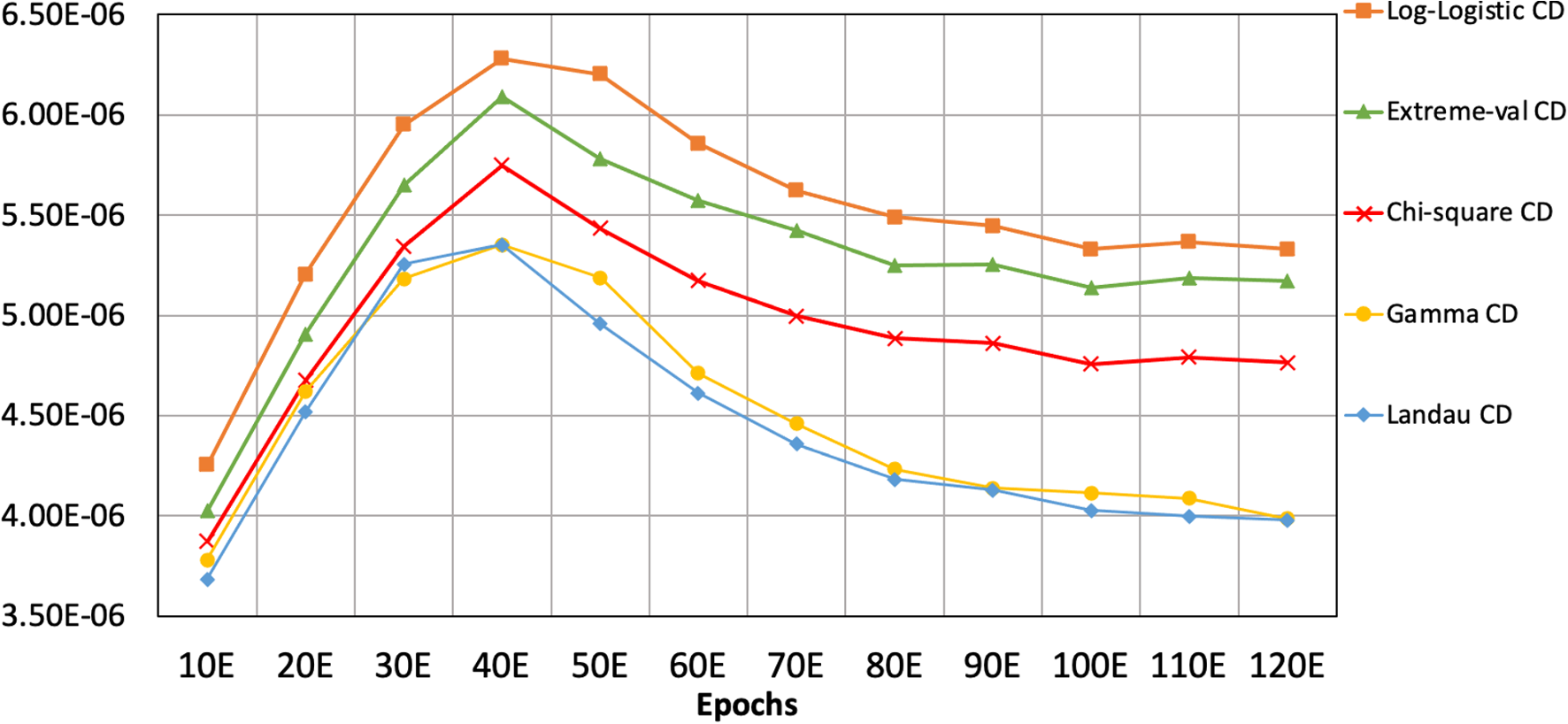}
	\end{center}
    \vspace{-5mm}
	\caption{Distances in network weights trained on different losses.}
	\label{fig:weights_diff}
    \vspace{-5mm}
\end{figure}

Our insight is to guide the learning of weighted loss functions with a well-known reference loss, as analogous to student/teacher network distillation. Fig. \ref{fig:weights_diff} illustrates the network weight distances during training on ShapeNet-Part with the same setting but different losses only. At the beginning, the weights are pushed away from HyperCD due to the error accumulation in gradients. However, after certain epochs the distances become smaller and smaller. Intuitively, this may be because all the network weights reach the wide and flat basin region of the loss landscape where HyperCD lies in, which explains why all of our weighted CD losses achieve similar results (a little higher or lower) to HyperCD.

\subsection{State-of-the-art Comparison}
\label{sec:Main_exp}

\begin{table*}[t]
    \centering	
    \small
	\setlength{\arrayrulewidth}{0.7pt}
	\caption{Comparison on PCN in terms of per-point L1-CD $\times 1000$ (lower is better).}
	\begin{tabular}{c|cccccccc|c}
		\toprule
		Methods & Plane & Cabinet & Car & Chair & Lamp & Couch & Table & Boat & Avg. \\
		\midrule
		FoldingNet \cite{yang2018foldingnet} & 14.31 & 9.49 & 15.80 & 12.61 & 15.55 & 16.41 & 15.97 & 13.65 & 14.99 \\
		TopNet \cite{topnet} & 7.61 & 13.31 & 10.90 & 13.82 & 14.44 & 14.78 & 11.22 & 11.12  & 12.15\\
 		AtlasNet \cite{groueix2018papier}  & 6.37 & 11.94 & 10.10 & 12.06 & 12.37 & 12.99 & 10.33 & 10.61 & 10.85\\
		GRNet \cite{GRNet}  & 6.45 & 10.37 & 9.45 & 9.41 & 7.96 & 10.51 & 8.44 & 8.04 & 8.83\\
		CRN \cite{wang2020cascaded}  & 4.79 & 9.97 & 8.31 & 9.49 & 8.94 & 10.69 & 7.81 & 8.05 & 8.51\\
		NSFA \cite{zhang2020detail}  & 4.76 & 10.18 & 8.63 & 8.53 & 7.03 & 10.53 & 7.35 & 7.48 & 8.06\\
		FBNet \cite{yan2022fbnet}  & 3.99 & 9.05 & 7.90 & 7.38 & 5.82 & 8.85 & 6.35 & 6.18 & 6.94\\
		PCN \cite{pcn}  & 5.50 & 22.70 & 10.63 & 8.70 & 11.00 & 11.34 & 11.68 & 8.59 & 11.27\\


        \hline 
        \hline
        
		FoldingNet \cite{yang2018foldingnet}  & 9.49 & 15.80 & 12.61 & 15.55 & 16.41 & 15.97 & 13.65 & 14.99 & 14.31\\

        HyperCD + FoldingNet & \textcolor{blue}{7.89} & 12.90 & \textcolor{blue}{10.67} & 14.55 & \textcolor{blue}{13.87} & \textcolor{blue}{14.09} & 11.86 & \textcolor{blue}{10.89}  & \textcolor{blue}{12.09}\\
       
        InfoCD + FoldingNet  & 7.90 & \textcolor{red}{12.68} & 10.83 & \textcolor{blue}{14.04} & 14.05 & 14.56 & \textcolor{blue}{11.61} & 11.45 & 12.14\\


        {\bf Landau CD + FoldingNet}  & \textcolor{red}{7.30} & \textcolor{blue}{12.69} & \textcolor{red}{10.46} & \textcolor{red}{13.00} & \textcolor{red}{11.92} & \textcolor{red}{13.39} & \textcolor{red}{10.86} & \textcolor{red}{10.59} & \textcolor{red}{11.27}\\

        \hline 
        \hline
        
		PMP-Net \cite{wen2021pmp}  & 5.65 & 11.24 & 9.64 & 9.51 & 6.95 & 10.83 & 8.72 & 7.25 & 8.73\\

        HyperCD + PMP-Net  & 5.06 & 10.67 & 9.30 & 9.11 & 6.83 & 11.01 & 8.18 & 7.03 & 8.40\\

        InfoCD + PMP-Net  & \textcolor{blue}{4.67} & \textcolor{red}{10.09} & \textcolor{red}{8.87} & \textcolor{blue}{8.59} & \textcolor{red}{6.38} & \textcolor{blue}{10.48} & \textcolor{blue}{7.51} & \textcolor{red}{6.75} & \textcolor{red}{7.92}\\

        {\bf Landau CD + PMP-Net}  & \textcolor{red}{4.59} & \textcolor{blue}{10.10} & \textcolor{blue}{8.90} & \textcolor{red}{8.57} & \textcolor{red}{6.38} & \textcolor{red}{10.47} & \textcolor{red}{7.49} & \textcolor{red}{6.75} & \textcolor{red}{7.92}\\

        \hline 
        \hline
        
		PoinTr \cite{pointtr}  & 4.75 & 10.47 & 8.68 & 9.39 & 7.75 & 10.93 & 7.78 & 7.29 & 8.38\\

        HyperCD + PoinTr  & 4.42 & 9.77 & 8.22 & 8.22 & 6.62 & 9.62 & 6.97 & 6.67 & 7.56\\

        InfoCD + PoinTr  & \textcolor{red}{4.06} & \textcolor{red}{9.42} & \textcolor{red}{8.11} & \textcolor{red}{7.81} & \textcolor{red}{6.21} & \textcolor{blue}{9.38} & \textcolor{red}{6.57} & \textcolor{red}{6.40} & \textcolor{red}{7.24}\\

         {\bf Landau CD + PoinTr}  & \textcolor{blue}{4.12} & \textcolor{blue}{9.49} & \textcolor{blue}{8.07} & \textcolor{blue}{7.82} & \textcolor{blue}{6.30} & \textcolor{red}{9.28} & \textcolor{blue}{6.76} & \textcolor{blue}{6.41} & \textcolor{blue}{7.28}\\

        \hline 
        \hline
        
		SnowflakeNet \cite{snowflakenet} & 4.29 & 9.16 & 8.08 & 7.89 & 6.07 & 9.23 & 6.55 & 6.40  & 7.21\\

        HyperCD + SnowflakeNet  & \textcolor{red}{3.95} & 9.01 & 7.88 & \textcolor{blue}{7.37} & \textcolor{red}{5.75} & 8.94 & \textcolor{blue}{6.19} & 6.17 & 6.91\\

        InfoCD + SnowflakeNet  & 4.01 & \textcolor{red}{8.81} & \textcolor{red}{7.62} & 7.51 & 5.80 & \textcolor{blue}{8.91} & 6.21 & \textcolor{red}{6.05} & \textcolor{red}{6.86}\\

        {\bf Landau CD + SnowflakeNet}  & \textcolor{blue}{3.98} & \textcolor{blue}{8.97} & \textcolor{blue}{7.78} & \textcolor{blue}{7.40} & \textcolor{blue}{5.76} & \textcolor{red}{8.86} & \textcolor{red}{6.16} & \textcolor{blue}{6.14} & \textcolor{blue}{6.88}\\

        \hline 
        \hline
        
		PointAttN \cite{wang2022pointattn}  & 3.87 & 9.00 & 7.63 & 7.43 & 5.90 & 8.68 & 6.32 & 6.09 & 6.86 \\

		DCD + PointAttN  & 4.47 & 9.65 & 8.14 & 8.12 & 6.75 & 9.60 & 6.92 & 6.67 & 7.54  \\

        HyperCD + PointAttN  & 3.76 & 8.93 & 7.49 & 7.06 & 5.61 & 8.48 & 6.25 & 5.92 & 6.68\\

         InfoCD + PointAttN   & \textcolor{red}{3.72} & \textcolor{red}{8.87} & \textcolor{red}{7.46} & \textcolor{red}{7.02} & \textcolor{red}{5.60} & \textcolor{red}{8.45} & \textcolor{red}{6.23} &\textcolor{red}{5.92} & \textcolor{red}{6.65}\\

        \midrule
        
        {\bf Gamma CD + PointAttN} & 3.83 & 8.96 & 7.58 & 7.15 & 5.69 & 8.56 & 6.34 & 6.01 & 6.76\\

        {\bf Chi-Squared CD + PointAttN}  & 3.77 & 8.93 & 7.49 & 7.08 & 5.64 & 8.50 & 6.28 & 5.95 & 6.70\\

        {\bf Log-Logistic CD + PointAttN}  & 3.78 & 8.92 & 7.47 & 7.10 & 5.63 & 8.51 & 6.29 & 5.94 & 6.70\\

        {\bf Extreme-Value CD + PointAttN}  & \textcolor{blue}{3.73} & \textcolor{blue}{8.88} & \textcolor{red}{7.46} & \textcolor{blue}{7.03} & \textcolor{blue}{5.61} & \textcolor{blue}{8.46} & 6.25 & \textcolor{red}{5.92} & \textcolor{blue}{6.66}\\
  
        {\bf Landau CD + PointAttN} & \textcolor{red}{3.72} & \textcolor{blue}{8.88} & \textcolor{red}{7.46} & 7.04 & \textcolor{red}{5.60} & 8.47 & \textcolor{blue}{6.24} & \textcolor{blue}{5.93} & \textcolor{blue}{6.66}\\
  
		\hline 
        \hline        

        SeedFormer \cite{zhou2022seedformer} & 3.85 & 9.05 & 8.06 & 7.06 & 5.21 & 8.85 & 6.05 & 5.85  & 6.74\\

        {DCD + SeedFormer}  & 16.42 & 26.23 & 21.08 & 20.06& 18.30 & 26.51 & 18.23 & 18.22 & 24.52\\             

        HyperCD + SeedFormer & 3.72 & \textcolor{blue}{8.71} & 7.79 & \textcolor{blue}{6.83} & 5.11 & 8.61 & 5.82 & 5.76 &6.54 \\

         InfoCD + SeedFormer  & \textcolor{blue}{3.69} & 8.72 & \textcolor{blue}{7.68} & 6.84 & \textcolor{blue}{5.08} & 8.61 & 5.83 & 5.75 & 6.52\\

        \midrule
        
         {\bf Log-Logistic CD + SeedFormer}  & 3.86 & 9.07 & 7.79 & 6.89 & 5.15 & 8.64 & 5.87 & 5.78 & 6.63\\

         {\bf Gamma CD + SeedFormer}  & 3.84 & 9.01 & 7.82 & 6.89 & 5.13 & 8.63 & 5.88 & 5.75 & 6.61\\
         
         {\bf Chi-Squared CD + SeedFormer}  & 3.75 & 8.90 & 7.71 & \textcolor{red}{6.80} & 5.11 & \textcolor{red}{8.48} & \textcolor{blue}{5.77} & 5.68 & 6.53\\
         

        {\bf Extreme-Value CD + SeedFormer} & 3.73 & 8.88 & 7.70 & \textcolor{red}{6.80} & \textcolor{blue}{5.08} & \textcolor{red}{8.48} & \textcolor{red}{5.75} & \textcolor{red}{5.65} & \textcolor{blue}{6.51}\\
         
        {\bf Landau CD + SeedFormer}  & \textcolor{blue}{3.65} & \textcolor{red}{8.68} & \textcolor{red}{7.64} & \textcolor{red}{6.80} & \textcolor{red}{5.04} & \textcolor{blue}{8.57} & 5.79 & \textcolor{blue}{5.71} & \textcolor{red}{6.49}\\

    \bottomrule
	\end{tabular}\vspace{3mm}
    \label{table:pcn}
\end{table*}

\begin{table*}[t]
\caption{Results on LiDAR scans from KITTI dataset under the  Fidelity and MMD metrics.} \small
\label{tab:KITTI}
\centering
\begin{tabular}[\linewidth]{l |c c c c|c c c c }
\toprule
& FoldingNet & HyperCD+F. & InfoCD+F.& {\bf Landau CD+F.} & PoinTr &  HyperCD+P. &  InfoCD+P.& {\bf Landau CD+P.} \\
\midrule
Fidelity $\downarrow$ & 7.467 &  2.214 & \textcolor{red}{1.944} & \textcolor{blue}{1.956} & \textcolor{red}{0.000} & \textcolor{red}{0.000} & \textcolor{red}{0.000}& \textcolor{red}{0.000} \\
MMD $\downarrow$ & 0.537 & 0.386  &  \textcolor{red}{0.333}&\textcolor{blue}{0.342}& 0.526& 0.507 & \textcolor{red}{0.502}&\textcolor{blue}{0.503}\\
\bottomrule
\end{tabular}
\end{table*}

\bfsection{PCN} In accordance with the literature, we report performances in terms of vanilla CD with L1-distance in Table~\ref{table:pcn}. As we can see, most of weighted CD losses achieve above-average results compared with the baseline networks used in training. In particular, we obtain some new state-of-the-art results using Landau CD. Meanwhile, Extreme-Value and Chi-Squared CD losses also achieve better performance than HyperCD in more complicated backbone networks PointAttN and SeedFormer. Also, InfoCD is a very strong baseline that performs similarly to Landau CD. In the sequel, by default we will only report the results using Landau CD, due to its great performance on PCN. 

\bfsection{KITTI}
In order to validate the effectiveness of weighted CD loss functions in real-world scenarios, we follow the method used in~\cite{GRNet} to fine tune two baseline models with Landau CD on ShapeNetCars~\cite{pcn} and evaluate the performance on KITTI. We report the Fidelity and MMD metrics in Table~\ref{tab:KITTI}. We observe that Landau CD can improve the baselines consistently with even better results compared with HyperCD. Furthermore, we present comprehensive visualization results in Fig.~\ref{fig:kitti}.
Note both HyperCD and Landau CD are able to recover the general geometrical structure from partial sparse input, Landau CD, however, perform better with less noise and outliers, especially on small details on corners and edges.

\begin{table*}[t]
	\centering	
	\small
	\setlength{\arrayrulewidth}{0.7pt}
	\caption{Results on ShapeNet-55 using L2-CD$\times 1000$ and F1 score.}
	\begin{tabular}{c|ccccc|cccc|c}
		\toprule
		Methods & Table & Chair & Plane & Car & Sofa & CD-S & CD-M & CD-H & Avg. & F1  \\
		\midrule

		FoldingNet   	& 2.53 & 2.81 & 1.43 & 1.98 & 2.48 & 2.67 & 2.66 & 4.05 & 3.12 & 0.082 \\

        HyperCD + FoldingNet & 2.35 & 2.62 & 1.25 &1.76  & 2.31 & 2.43 & \textcolor{red}{2.45} &3.88  & 2.92 & 0.109 \\
  
  	InfoCD + FoldingNet & \textcolor{blue}{2.14} & \textcolor{blue}{2.37} & \textcolor{blue}{1.03} & \textcolor{blue}{1.55} & \textcolor{blue}{2.04} & \textcolor{blue}{2.17} & 2.50 & \textcolor{blue}{3.46} & \textcolor{blue}{2.71} & \textcolor{blue}{0.137} \\
        
        {\bf Landau CD + FoldingNet} & \textcolor{red}{2.09} & \textcolor{red}{2.32} & \textcolor{red}{1.01} & \textcolor{red}{1.50} & \textcolor{red}{2.01} & \textcolor{red}{2.15} & \textcolor{blue}{2.46} & \textcolor{red}{3.39} & \textcolor{red}{2.66} & \textcolor{red}{0.141} \\
		\midrule

        PoinTr  			& 0.81 & 0.95 & 0.44 & 0.91 & 0.79 & 0.58 & 0.88 & 1.79 & 1.09 & 0.464 \\

        HyperCD + PoinTr 	& 0.79 & 0.92 & 0.41 & 0.90 & 0.76 &0.54  &0.85  & 1.73 & 1.04 & 0.499 \\

		InfoCD + PoinTr & \textcolor{blue}{0.69} & \textcolor{blue}{0.83} & \textcolor{blue}{0.33} & \textcolor{blue}{0.80} & \textcolor{blue}{0.67} & \textcolor{blue}{0.47} & \textcolor{blue}{0.73} & \textcolor{blue}{1.50} & \textcolor{blue}{0.90} & \textcolor{blue}{0.524} \\
  
		{\bf Landau CD + PoinTr} & \textcolor{red}{0.63} & \textcolor{red}{0.82} & \textcolor{red}{0.32} & \textcolor{red}{0.78} & \textcolor{red}{0.65} & \textcolor{red}{0.43} & \textcolor{red}{0.70} & \textcolor{red}{1.47} & \textcolor{red}{0.88} & \textcolor{red}{0.527} \\
		\midrule
        
        SeedFormer  & 0.72 & 0.81 & 0.40 & 0.89 & 0.71 & 0.50 & 0.77 & 1.49 & 0.92 & 0.472 \\       
		HyperCD + SeedFormer & \textcolor{blue}{0.66} & 0.74 & 0.35 & 0.83 & \textcolor{blue}{0.64} & 0.47 & \textcolor{blue}{0.72} & 1.40 & \textcolor{blue}{0.86} & 0.482 \\

		InfoCD + SeedFormer & \textcolor{red}{0.65} & \textcolor{red}{0.72} & \textcolor{red}{0.31} & \textcolor{red}{0.81} & \textcolor{red}{0.62} & \textcolor{red}{0.43} & \textcolor{red}{0.71} & \textcolor{red}{1.38} & \textcolor{red}{0.84} & \textcolor{red}{0.490} \\

		{\bf Landau CD + SeedFormer} & 0.67 & \textcolor{blue}{0.73} & \textcolor{blue}{0.34} & \textcolor{blue}{0.82} & \textcolor{red}{0.62} & \textcolor{blue}{0.45} & 0.73 & \textcolor{blue}{1.39} & \textcolor{blue}{0.86} & \textcolor{blue}{0.489} \\
		\bottomrule
	\end{tabular}

	\label{table:shapenet55}
\end{table*}

\begin{table*}[t]
	\centering
	\small
	\setlength{\arrayrulewidth}{0.7pt}
	\caption{Results on ShapeNet-34 using L2-CD$\times 1000$ and F1 score.}
	\begin{tabular}{c|ccccc|ccccc}
		\toprule
		\multirow{2}{*}{Methods} & \multicolumn{5}{c|}{34 seen categories} & \multicolumn{5}{c}{21 unseen categories} \\
		        & CD-S & CD-M & CD-H & Avg. & F1 & CD-S & CD-M & CD-H & Avg. & F1 \\
		\midrule
        FoldingNet   	 & 1.86 & 1.81 & 3.38 & 2.35 & 0.139 & 2.76 & 2.74 & 5.36 & 3.62 & 0.095 \\

        HyperCD + FoldingNet  	 & 1.71 & 1.69 & 3.23 & 2.21 & 0.148 & 2.55 & 2.59 & 5.19 & 3.44 & 0.122 \\

 	InfoCD + FoldingNet & \textcolor{blue}{1.54} & \textcolor{blue}{1.60} & \textcolor{blue}{3.10} & \textcolor{blue}{2.08} & \textcolor{blue}{0.177}  & \textcolor{blue}{2.42} & \textcolor{blue}{2.49} & \textcolor{red}{5.01} & \textcolor{blue}{3.31} & \textcolor{blue}{0.157}\\

 		{\bf Landau CD + FoldingNet} & \textcolor{red}{1.50} & \textcolor{red}{1.57} & \textcolor{red}{3.04} & \textcolor{red}{2.03} & \textcolor{red}{0.179}  & \textcolor{red}{2.40} & \textcolor{red}{2.45} & \textcolor{blue}{5.02} & \textcolor{red}{3.29} & \textcolor{red}{0.158}\\
        \midrule
        
        PoinTr    			 & 0.76 & 1.05 & 1.88 & 1.23 & 0.421 & 1.04 & 1.67 & 3.44 & 2.05 & 0.384 \\

        HyperCD + PoinTr   			 & 0.72 & 1.01 & 1.85 & 1.19 & 0.428 & 1.01 & 1.63 & 3.40 & \textcolor{blue}{2.01} & 0.389 \\

		InfoCD + PoinTr & \textcolor{blue}{0.47} & \textcolor{blue}{0.69} & \textcolor{blue}{1.35} & \textcolor{blue}{0.84} & \textcolor{blue}{0.529} & \textcolor{red}{0.61} & \textcolor{blue}{1.06} & \textcolor{blue}{2.55} & \textcolor{red}{1.41} & \textcolor{red}{0.493} \\
        
		{\bf Landau CD + PoinTr} & \textcolor{red}{0.44} & \textcolor{red}{0.65} & \textcolor{red}{1.29} & \textcolor{red}{0.79} & \textcolor{red}{0.531} & \textcolor{blue}{0.63} & \textcolor{red}{1.07} & \textcolor{red}{2.54} & \textcolor{red}{1.41} & \textcolor{blue}{0.492} \\
        \midrule
        SeedFormer     & 0.48 & 0.70 & 1.30 & 0.83 & 0.452 & 0.61 & 1.08 & 2.37 & 1.35 & 0.402 \\ 
        
		HyperCD + SeedFormer & 0.46 & 0.67 & 1.24 & 0.79 & 0.459 & 0.58 & 1.03 & 2.24 & 1.31 & 0.428 \\

		InfoCD + SeedFormer & \textcolor{blue}{0.43} & \textcolor{red}{0.63} & \textcolor{red}{1.21} & \textcolor{red}{0.75} & \textcolor{red}{0.581} & \textcolor{red}{0.54} & \textcolor{red}{1.01} & \textcolor{blue}{2.18} & \textcolor{red}{1.24} & \textcolor{red}{0.449} \\
   
		{\bf Landau CD + SeedFormer} & \textcolor{red}{0.42} & \textcolor{blue}{0.64} & \textcolor{blue}{1.23} & \textcolor{blue}{0.76} & \textcolor{blue}{0.580} & \textcolor{blue}{0.56} & \textcolor{blue}{1.03} & \textcolor{red}{2.17} & \textcolor{blue}{1.25} & \textcolor{blue}{0.447} \\
		\bottomrule
	\end{tabular}

	\label{table:shapenet34}
\end{table*}

\bfsection{ShapeNet-55/34}
To assess the adaptability of Landau CD, we evaluate it on the datasets with higher diversities. 
\begin{itemize}[nosep, leftmargin=*]
    \item {\em ShapeNet-55:} Table~\ref{table:shapenet55} enumerates the performance across three levels of difficulty with the average. Qualitative evaluation results are shown in Fig.~\ref{fig:shapnet55} as well from Seedformer trained with HyperCD and Landau CD as a supplement to numerical values. Landau CD exhibits comparable but better performance than HyperCD, which matches our intuition. Visually, Landau CD also shows slightly better smoothing on surfaces than HyperCD (see the bases of lamps in both outputs).

    \item {\em ShapeNet-34:} We assess performances within 34 seen categories (used in training) as well as 21 unseen categories (not used in training), and list our results in Table~\ref{table:shapenet34}. We can see that, once again, Landau CD improves the performance of baseline models, suggesting that Landau CD, as a representative of our weighted CD loss family, is highly generalizable for point cloud completion. 
\end{itemize}



\bfsection{Single View Reconstruction} We extend our evaluation to {\em Single View Reconstruction (SVR)} task aims to reconstruct a point cloud from a single image of the target object. Following the literature such as \cite{wen20223d, snowflakenet}, we sample 30K points from the watertight mesh in ShapeNet as the ground truth and output 2048 points for evaluation based on per-point L1-CD$\times10^2$. We replace the CD loss in SnowflakeNet \cite{snowflakenet} with different losses in training. Comparative results are presented in Table \ref{tab:svr}, showing that weighted CD losses can demonstrate comparable performance to HyperCD and InfoCD in SVR.
\begin{wraptable}{r}{.5\linewidth}
\vspace{-2mm}
\setlength{\intextsep}{0pt}
\setlength{\columnsep}{0pt}
    \caption{{SVR results.}}\label{tab:svr}
    \setlength{\tabcolsep}{1pt}
    \small
        \begin{tabular}{c|c}
            \toprule
            Method & Ave. \\
            \midrule
            3DAttriFlow \cite{wen20223d}  & 3.02 \\
            SnowflakeNet \cite{snowflakenet} &2.86 \\
            \midrule
             HyperCD + SnowflakeNet & \textcolor{red}{ 2.73} \\
            InfoCD + SnowflakeNet & \textcolor{red}{ 2.73} \\            
           {\bf  Chi-Squared CD + S.} & 2.75 \\ 
            {\bf Extreme-Value CD + S.} & \textcolor{blue}{2.74} \\ 
            {\bf Landau CD + S.} & \textcolor{red}{2.73} \\ 
            \bottomrule
    	\end{tabular}    
     \vspace{-4mm}
\end{wraptable}

%% file: sections/5_conclusion.tex
\section{Conclusion}
\label{sec:con}
In this paper, we propose a novel loss distillation method for point cloud completion by mimicking the learning behavior of HyperCD based on weighted CD. To this end, we propose an efficient and effective gradient matching algorithm to search for potential weighting functions from a pool of distributions for weighted CD by comparing them with the reference curve from HyperCD. This eventually converts to a bilevel optimization problem in training backbone networks, with empirical convergence based on our iterative differentiation algorithm. We conduct comprehensive experiments using real-world datasets such as KITTI\cite{geiger2013vision} to demonstrate the effectiveness of weighted CD losses, particularly Landau CD which achieves new state-of-the-art results on several benchmark datasets.

\bfsection{Limitations}
A pool of candidate weighting functions needs to be defined first for gradient matching, which may need prior knowledge of the target function (\eg HyperCD). Currently, we do not have an efficient way to determine which weighting functions will certainly improve the performance, and this has to be tested empirically. Also, our loss parameters are simply chosen by the gradient matching, which may limit the potentials of weighted CD. In our future work, we plan to address these issues.

%% file: main.bbl
\begin{thebibliography}{10}
\providecommand{\url}[1]{#1}
\csname url@rmstyle\endcsname
\providecommand{\newblock}{\relax}
\providecommand{\bibinfo}[2]{#2}
\providecommand\BIBentrySTDinterwordspacing{\spaceskip=0pt\relax}
\providecommand\BIBentryALTinterwordstretchfactor{4}
\providecommand\BIBentryALTinterwordspacing{\spaceskip=\fontdimen2\font plus
\BIBentryALTinterwordstretchfactor\fontdimen3\font minus \fontdimen4\font\relax}
\providecommand\BIBforeignlanguage[2]{{%
\expandafter\ifx\csname l@#1\endcsname\relax
\typeout{** WARNING: IEEEtran.bst: No hyphenation pattern has been}%
\typeout{** loaded for the language `#1'. Using the pattern for}%
\typeout{** the default language instead.}%
\else
\language=\csname l@#1\endcsname
\fi
#2}}

\bibitem{lin2023hyperbolic}
F.~Lin, Y.~Yue, S.~Hou, X.~Yu, Y.~Xu, D.~K. Yamada, and Z.~Zhang, ``Hyperbolic chamfer distance for point cloud completion,'' in \emph{Proceedings of the IEEE/CVF international conference on computer vision}, 2023.

\bibitem{xie2018attentional}
S.~Xie, S.~Liu, Z.~Chen, and Z.~Tu, ``Attentional shapecontextnet for point cloud recognition,'' in \emph{Proceedings of the IEEE conference on computer vision and pattern recognition}, 2018, pp. 4606--4615.

\bibitem{XU2022255}
\BIBentryALTinterwordspacing
Y.~Xu, S.~Arai, D.~Liu, F.~Lin, and K.~Kosuge, ``Fpcc: Fast point cloud clustering-based instance segmentation for industrial bin-picking,'' \emph{Neurocomputing}, vol. 494, pp. 255--268, 2022. [Online]. Available: \url{https://www.sciencedirect.com/science/article/pii/S0925231222003915}
\BIBentrySTDinterwordspacing

\bibitem{huitl2012tumindoor}
R.~Huitl, G.~Schroth, S.~Hilsenbeck, F.~Schweiger, and E.~Steinbach, ``Tumindoor: An extensive image and point cloud dataset for visual indoor localization and mapping,'' in \emph{2012 19th IEEE International Conference on Image Processing}.\hskip 1em plus 0.5em minus 0.4em\relax IEEE, 2012, pp. 1773--1776.

\bibitem{hu2024orbitgrasp}
B.~Hu, X.~Zhu, D.~Wang, Z.~Dong, H.~Huang, C.~Wang, R.~Walters, and R.~Platt, ``Orbitgrasp: {SE}(3)-equivariant grasp learning,'' \emph{arXiv preprint arXiv:2407.03531}, 2024.

\bibitem{huang2023edge}
H.~Huang, D.~Wang, X.~Zhu, R.~Walters, and R.~Platt, ``Edge grasp network: A graph-based se (3)-invariant approach to grasp detection,'' in \emph{2023 IEEE International Conference on Robotics and Automation (ICRA)}.\hskip 1em plus 0.5em minus 0.4em\relax IEEE, 2023, pp. 3882--3888.

\bibitem{huang2024imagination}
H.~Huang, K.~Schmeckpeper, D.~Wang, O.~Biza, Y.~Qian, H.~Liu, M.~Jia, R.~Platt, and R.~Walters, ``Imagination policy: Using generative point cloud models for learning manipulation policies,'' \emph{arXiv preprint arXiv:2406.11740}, 2024.

\bibitem{ten2017grasp}
A.~Ten~Pas, M.~Gualtieri, K.~Saenko, and R.~Platt, ``Grasp pose detection in point clouds,'' \emph{The International Journal of Robotics Research}, vol.~36, no. 13-14, pp. 1455--1473, 2017.

\bibitem{leberl2010point}
F.~Leberl, A.~Irschara, T.~Pock, P.~Meixner, M.~Gruber, S.~Scholz, and A.~Wiechert, ``Point clouds,'' \emph{Photogrammetric Engineering \& Remote Sensing}, vol.~76, no.~10, pp. 1123--1134, 2010.

\bibitem{guo2020deep}
Y.~Guo, H.~Wang, Q.~Hu, H.~Liu, L.~Liu, and M.~Bennamoun, ``Deep learning for 3d point clouds: A survey,'' \emph{IEEE transactions on pattern analysis and machine intelligence}, vol.~43, no.~12, pp. 4338--4364, 2020.

\bibitem{pointtr}
X.~Yu, Y.~Rao, Z.~Wang, Z.~Liu, J.~Lu, and J.~Zhou, ``Pointr: Diverse point cloud completion with geometry-aware transformers,'' in \emph{ICCV}, 2021.

\bibitem{snowflakenet}
P.~Xiang, X.~Wen, Y.-S. Liu, Y.-P. Cao, P.~Wan, W.~Zheng, and Z.~Han, ``Snowflakenet: Point cloud completion by snowflake point deconvolution with skip-transformer,'' in \emph{ICCV}, 2021.

\bibitem{zhou2022seedformer}
H.~Zhou, Y.~Cao, W.~Chu, J.~Zhu, T.~Lu, Y.~Tai, and C.~Wang, ``Seedformer: Patch seeds based point cloud completion with upsample transformer,'' \emph{arXiv preprint arXiv:2207.10315}, 2022.

\bibitem{wang2022pointattn}
J.~Wang, Y.~Cui, D.~Guo, J.~Li, Q.~Liu, and C.~Shen, ``Pointattn: You only need attention for point cloud completion,'' \emph{arXiv preprint arXiv:2203.08485}, 2022.

\bibitem{wu2021densityaware}
T.~Wu, L.~Pan, J.~Zhang, T.~Wang, Z.~Liu, and D.~Lin, ``Density-aware chamfer distance as a comprehensive metric for point cloud completion,'' in \emph{Advances in Neural Information Processing Systems}, vol.~34, 2021, pp. 29\,088--29\,100.

\bibitem{lin2023infocd}
F.~Lin, Y.~Yue, Z.~Zhang, S.~Hou, K.~Yamada, V.~B. Kolachalama, and V.~Saligrama, ``Info{CD}: A contrastive chamfer distance loss for point cloud completion,'' in \emph{Thirty-seventh Conference on Neural Information Processing Systems}, 2023.

\bibitem{dis_pc}
F.~M{\'e}moli and G.~Sapiro, ``Comparing point clouds,'' in \emph{Proceedings of the 2004 Eurographics/ACM SIGGRAPH symposium on Geometry processing}, 2004, pp. 32--40.

\bibitem{ravi2020pytorch3d}
N.~Ravi, J.~Reizenstein, D.~Novotny, T.~Gordon, W.-Y. Lo, J.~Johnson, and G.~Gkioxari, ``Accelerating 3d deep learning with pytorch3d,'' \emph{arXiv:2007.08501}, 2020.

\bibitem{deng20193d}
H.~Deng, T.~Birdal, and S.~Ilic, ``3d local features for direct pairwise registration,'' in \emph{Proceedings of the IEEE/CVF Conference on Computer Vision and Pattern Recognition}, 2019, pp. 3244--3253.

\bibitem{lyu2021conditional}
Z.~Lyu, Z.~Kong, X.~Xu, L.~Pan, and D.~Lin, ``A conditional point diffusion-refinement paradigm for 3d point cloud completion,'' \emph{arXiv preprint arXiv:2112.03530}, 2021.

\bibitem{zhang2022attention}
K.~Zhang, X.~Yang, Y.~Wu, and C.~Jin, ``Attention-based transformation from latent features to point clouds,'' in \emph{Proceedings of the AAAI Conference on Artificial Intelligence}, vol.~36, no.~3, 2022, pp. 3291--3299.

\bibitem{tang2022lake}
J.~Tang, Z.~Gong, R.~Yi, Y.~Xie, and L.~Ma, ``Lake-net: topology-aware point cloud completion by localizing aligned keypoints,'' in \emph{Proceedings of the IEEE/CVF conference on computer vision and pattern recognition}, 2022, pp. 1726--1735.

\bibitem{knowledge_distilling}
G.~Hinton, O.~Vinyals, and J.~Dean, ``Distilling the knowledge in a neural network,'' \emph{arXiv preprint arXiv:1503.02531}, 2015.

\bibitem{kd_teacher_student}
L.~Wang and K.-J. Yoon, ``Knowledge distillation and student-teacher learning for visual intelligence: A review and new outlooks,'' \emph{IEEE transactions on pattern analysis and machine intelligence}, vol.~44, no.~6, pp. 3048--3068, 2021.

\bibitem{gou2021knowledge}
J.~Gou, B.~Yu, S.~J. Maybank, and D.~Tao, ``Knowledge distillation: A survey,'' \emph{International Journal of Computer Vision}, vol. 129, pp. 1789--1819, 2021.

\bibitem{wang2021knowledge}
L.~Wang and K.-J. Yoon, ``Knowledge distillation and student-teacher learning for visual intelligence: A review and new outlooks,'' \emph{IEEE transactions on pattern analysis and machine intelligence}, vol.~44, no.~6, pp. 3048--3068, 2021.

\bibitem{alkhulaifi2021knowledge}
A.~Alkhulaifi, F.~Alsahli, and I.~Ahmad, ``Knowledge distillation in deep learning and its applications,'' \emph{PeerJ Computer Science}, vol.~7, p. e474, 2021.

\bibitem{WD_PointLattice}
C.~Fouard, R.~Strand, and G.~Borgefors, ``Weighted distance transforms generalized to modules and their computation on point lattices,'' \emph{Pattern Recognition}, vol.~40, no.~9, pp. 2453--2474, 2007.

\bibitem{ImageSeg_via_WD}
A.~Protiere and G.~Sapiro, ``Interactive image segmentation via adaptive weighted distances,'' \emph{IEEE Transactions on Image Processing}, vol.~16, no.~4, pp. 1046--1057, 2007.

\bibitem{fcc}
K.~Petkov, F.~Qiu, Z.~Fan, A.~E. Kaufman, and K.~Mueller, ``Efficient lbm visual simulation on face-centered cubic lattices,'' \emph{IEEE Transactions on Visualization and Computer Graphics}, vol.~15, no.~5, pp. 802--814, 2009.

\bibitem{bcc}
A.~Entezari, D.~Van De~Ville, and T.~Moller, ``Practical box splines for reconstruction on the body centered cubic lattice,'' \emph{IEEE Transactions on Visualization and Computer Graphics}, vol.~14, no.~2, pp. 313--328, 2008.

\bibitem{loss_object_func}
M.~A. Gennert and A.~L. Yuille, ``Determining the optimal weights in multiple objective function optimization,'' in \emph{ICCV}, 1988, pp. 87--89.

\bibitem{neighbor_loss}
A.~F. Guarda, N.~M. Rodrigues, and F.~Pereira, ``Neighborhood adaptive loss function for deep learning-based point cloud coding with implicit and explicit quantization,'' \emph{IEEE MultiMedia}, vol.~28, no.~3, pp. 107--116, 2020.

\bibitem{deep_p_dis}
D.~Urbach, Y.~Ben-Shabat, and M.~Lindenbaum, ``Dpdist: Comparing point clouds using deep point cloud distance,'' in \emph{Computer Vision--ECCV 2020: 16th European Conference, Glasgow, UK, August 23--28, 2020, Proceedings, Part XI 16}.\hskip 1em plus 0.5em minus 0.4em\relax Springer, 2020, pp. 545--560.

\bibitem{ji2021bilevel}
K.~Ji, J.~Yang, and Y.~Liang, ``Bilevel optimization: Convergence analysis and enhanced design,'' in \emph{International conference on machine learning}.\hskip 1em plus 0.5em minus 0.4em\relax PMLR, 2021, pp. 4882--4892.

\bibitem{pcn}
W.~Yuan, T.~Khot, D.~Held, C.~Mertz, and M.~Hebert, ``Pcn: point completion network,'' in \emph{3DV}, 2018.

\bibitem{yi2016scalable}
L.~Yi, V.~G. Kim, D.~Ceylan, I.-C. Shen, M.~Yan, H.~Su, C.~Lu, Q.~Huang, A.~Sheffer, and L.~Guibas, ``A scalable active framework for region annotation in 3d shape collections,'' \emph{ACM Transactions on Graphics (ToG)}, vol.~35, no.~6, pp. 1--12, 2016.

\bibitem{geiger2013vision}
A.~Geiger, P.~Lenz, C.~Stiller, and R.~Urtasun, ``Vision meets robotics: The kitti dataset,'' \emph{The International Journal of Robotics Research}, vol.~32, no.~11, pp. 1231--1237, 2013.

\bibitem{xiang2022snowflake}
P.~Xiang, X.~Wen, Y.-S. Liu, Y.-P. Cao, P.~Wan, W.~Zheng, and Z.~Han, ``Snowflake point deconvolution for point cloud completion and generation with skip-transformer,'' \emph{IEEE Transactions on Pattern Analysis and Machine Intelligence}, vol.~45, no.~5, pp. 6320--6338, 2022.

\bibitem{chang2015shapenet}
A.~X. Chang, T.~Funkhouser, L.~Guibas, P.~Hanrahan, Q.~Huang, Z.~Li, S.~Savarese, M.~Savva, S.~Song, H.~Su, \emph{et~al.}, ``Shapenet: An information-rich 3d model repository,'' \emph{arXiv preprint arXiv:1512.03012}, 2015.

\bibitem{yang2018foldingnet}
Y.~Yang, C.~Feng, Y.~Shen, and D.~Tian, ``Foldingnet: Point cloud auto-encoder via deep grid deformation,'' in \emph{Proceedings of the IEEE conference on computer vision and pattern recognition}, 2018, pp. 206--215.

\bibitem{wen2021pmp}
X.~Wen, P.~Xiang, Z.~Han, Y.-P. Cao, P.~Wan, W.~Zheng, and Y.-S. Liu, ``Pmp-net: Point cloud completion by learning multi-step point moving paths,'' in \emph{Proceedings of the IEEE/CVF Conference on Computer Vision and Pattern Recognition}, 2021, pp. 7443--7452.

\bibitem{lin2022cosmos}
F.~Lin, Y.~Xu, Z.~Zhang, C.~Gao, and K.~D. Yamada, ``Cosmos propagation network: Deep learning model for point cloud completion,'' \emph{Neurocomputing}, vol. 507, pp. 221--234, 2022.

\bibitem{tatarchenko2019single}
M.~Tatarchenko, S.~R. Richter, R.~Ranftl, Z.~Li, V.~Koltun, and T.~Brox, ``What do single-view 3d reconstruction networks learn?'' in \emph{Proceedings of the IEEE/CVF Conference on Computer Vision and Pattern Recognition}, 2019, pp. 3405--3414.

\bibitem{topnet}
L.~P. Tchapmi, V.~Kosaraju, H.~Rezatofighi, I.~Reid, and S.~Savarese, ``Topnet: Structural point cloud decoder,'' in \emph{CVPR}, 2019.

\bibitem{groueix2018papier}
T.~Groueix, M.~Fisher, V.~G. Kim, B.~C. Russell, and M.~Aubry, ``A papier-m{\^a}ch{\'e} approach to learning 3d surface generation,'' in \emph{Proceedings of the IEEE conference on computer vision and pattern recognition}, 2018, pp. 216--224.

\bibitem{GRNet}
H.~Xie, H.~Yao, S.~Zhou, J.~Mao, S.~Zhang, and W.~Sun, ``Grnet: Gridding residual network for dense point cloud completion,'' in \emph{ECCV}, 2020.

\bibitem{wang2020cascaded}
X.~Wang, M.~H. Ang~Jr, and G.~H. Lee, ``Cascaded refinement network for point cloud completion,'' in \emph{Proceedings of the IEEE/CVF Conference on Computer Vision and Pattern Recognition}, 2020, pp. 790--799.

\bibitem{zhang2020detail}
W.~Zhang, Q.~Yan, and C.~Xiao, ``Detail preserved point cloud completion via separated feature aggregation,'' in \emph{European Conference on Computer Vision}.\hskip 1em plus 0.5em minus 0.4em\relax Springer, 2020, pp. 512--528.

\bibitem{yan2022fbnet}
X.~Yan, H.~Yan, J.~Wang, H.~Du, Z.~Wu, D.~Xie, S.~Pu, and L.~Lu, ``Fbnet: Feedback network for point cloud completion,'' in \emph{European Conference on Computer Vision}.\hskip 1em plus 0.5em minus 0.4em\relax Springer, 2022, pp. 676--693.

\bibitem{wen20223d}
X.~Wen, J.~Zhou, Y.-S. Liu, H.~Su, Z.~Dong, and Z.~Han, ``3d shape reconstruction from 2d images with disentangled attribute flow,'' in \emph{Proceedings of the IEEE/CVF conference on computer vision and pattern recognition}, 2022, pp. 3803--3813.

\end{thebibliography}
